\documentclass{article}
\usepackage{amsmath}
\usepackage{amsfonts} 
\usepackage{graphicx} 
\usepackage{adjustbox} 

\usepackage{caption}
\usepackage{subcaption}
\usepackage{booktabs}
\usepackage{changepage}
\usepackage{xcolor}
\usepackage{bm}
\usepackage{mathrsfs}
\usepackage{array}
\usepackage[authoryear, round]{natbib}

\usepackage[normalem]{ulem}

\newcommand{\BR}{\mathbb{R}}

\newcommand{\BP}{\mathbb{P}}

\newcommand{\CL}{\mathcal{L}}

\usepackage{algorithm}
\usepackage{etoolbox} 
\let\classAND\AND
\let\AND\relax
\usepackage{algorithmic}

\let\AND\classAND
\AtBeginEnvironment{algorithmic}{\let\AND\algoAND}
\usepackage{bm}

\title{Self-supervised Representation Learning for Cell Event Recognition through Time Arrow Prediction}

\author{
  {
  Cangxiong Chen
  \thanks{Institute for Mathematical Innovation, University of Bath}
  } 
  {Vinay P.~Namboodiri
  \thanks{Department of Computer Science, University of Bath}
  }
  {
  Julia E. Sero
  \thanks{
  Department of Life Sciences, University of Bath
  }
  }
}
\date{5th November 2024}
\begin{document}
\maketitle
\begin{abstract}
    The spatio-temporal nature of live-cell microscopy data poses challenges in the analysis of cell states which is fundamental in bioimaging. Deep-learning based segmentation or tracking methods rely on large amount of high quality annotations to work effectively. In this work, we explore an alternative solution: using feature maps obtained from self-supervised representation learning (SSRL) on time arrow prediction (TAP) for the downstream supervised task of cell event recognition. We demonstrate through extensive experiments and analysis that this approach can achieve better performance with limited annotation compared to models trained from end to end using fully supervised approach. Our analysis also provides insight into applications of the SSRL using TAP in live-cell microscopy.  
    
\end{abstract}
\section{Introduction}
In live-cell microscopy where the data is often spatio-temporal by nature, one interesting problem is recognising cell divisions and deaths. Most existing methods using machine learning focus on cell segmentation and tracking (\cite{GreenwaldWholeCell2022}, \cite{StringerCellpose2020}, \cite{WeigertStarConvex2020}, \cite{WeigertCellDetectionwithStar2018}). This often requires large amount of annotated data which can be costly to obtain. A new paradigm which appeared recently is to leverage self-supervised representation learning (SSRL) \citep{EricssonSelfSupervisedRepresentation2022} which is capable of building feature maps from pretext tasks trained on unlabelled data (e.g. frames from movies of live cells) and apply the obtained dense features in downstream tasks such as cell event recognition or cell tracking. By incorporating SSRL, we are able to boost the performance of the model on downstream tasks using the same amount of annotated data compared to models trained from end to end using fully supervised learning. The benefits of SSRL become more evident with spatio-temporal data compared with static cell images, where human annotations become even more time consuming due to the extra effort needed to annotate each frame over time. In this paper, we built on the work from \cite{GallusserSelfsupervisedDense2023} and apply SSRL to the problem of cell event recognition where for a given image pair following the time direction, we predict whether there are cell divisions and deaths happening. Compared to \cite{GallusserSelfsupervisedDense2023}, our work provides a comprehensive evaluation of cell event recognition using features maps learned from SSRL and offers insight into the application of SSRL for cell event recognition.

\paragraph{Our contributions} 
\begin{enumerate}
    \item [1] We demonstrate through extensive experiments that using time arrow prediction (TAP) \citep{PickupSeeingtheArrow2014} is a very effective way to obtain dense features that is extremely useful for the downstream task of cell event recognition. When we apply the feature map trained from TAP with fine-tuning on our annotated dataset, we achieve better performance compared to a fully supervised approach from end to end. 
    \item [2] We provide analysis on the mistaken predictions, carry out comparisons on the performance from various labelling criteria and calibrate the predictions using temperature scaling \citep{Guo2017OnCalibration}. We hope these analyses can provide insights to the application of SSRL using TAP in live-cell microscopy.
\end{enumerate}

In Section \ref{sec:cell_event_recog}, we give a formal description of our learning problem and methods. In Section \ref{sec:experiments}, we explain the details of our data pre-processing method, followed by results on evaluation of TAP features, analysis of mistaken predictions, comparisons on labelling criteria and results on model calibration. We review related work in Section \ref{sec:related_work} and summarise limitations and future work in Section \ref{sec:limit_future_work}.

\section{Method}\label{sec:cell_event_recog}
One of the key problems in representation learning is learning good features that can benefit downstream tasks that we are interested in. Inspired by \cite{GallusserSelfsupervisedDense2023}, we will learn a set of dense features from predicting the time arrow, i.e. which of the order of occurrence for a pair of image patches at the same spatial location. We believe that these features will be crucial for the downstream task of cell event recognition because events such as division and death have strong time arrow property. 

\paragraph{Learning features from time arrow predictions}
Let $\{x_t\}, t \in [0,T], x_t \in \BR^{h \times w}$ be s sequence of images indexed by time $t$. The goal of time arrow prediction is to learn a feature map $f \colon \BR^{h \times w} \rightarrow \BR^{c_0 \times h_0 \times w_0}$ so that the dense feature $z_t := f(x_t)$ given by $f$ can be used for the subsequent task of predicting the direction of time for any given image pair at the same spatial location. This can be achieved by minimising the binary cross-entropy loss with respect to $f$ and the classification head $h$. In addition, it is helpful to consider a regularisation term in the loss that penalises the correlation among feature channels of $z_t$ \citep{GallusserSelfsupervisedDense2023}. Details of the loss terms can be found in the Appendix.

\paragraph{Cell event recognition}
Now we describe in detail of our formulation of the downstream task of cell event recognition. Let $f$ denote the feature map learned from time arrow prediction (TAP). For image patches $(x_t, x_{t+\Delta t})$, we denote the corresponding dense features by $z_t := f(x_t), z_{t + \Delta t} := f(x_{t+\Delta t}).$ We are interested in identifying whether there is death or division happening from $x_t$ to $x_{t+\Delta t}$. More precisely, we are interested in the probability of the event $E$ that either death or division happens in either $x_t$ or $x_{t+\Delta t}$ conditioning on the features $z_t$ and $z_{t + \Delta t}$:
\begin{equation}\label{celleventrecogformulation}
    \hat{y} := \BP( E | z_t, z_{t + \Delta t}).
\end{equation}
Here $\hat{y}$ denotes the likelihood of the event of death or division happens in either of the patches $x_t$ or $x_{t+\Delta t}$. Notice that unlike TAP, cell event recognition is time invariant. That is, flipping the order of $x_t$ $x_{t+\Delta t}$ (and so of $z_t, z_{t + \Delta t}$) should not change the probability \eqref{celleventrecogformulation}.

To train a model for cell event recognition, we fix the feature map $f$ and only train the prediction head $h$:
\begin{equation}
    h: (z_t, z_{t + \Delta t};w) \mapsto (\hat{y}_1, \hat{y}_2)
\end{equation}
by minimising the cross entropy function: 
\begin{equation}
\CL(z_t, z_{t + \Delta t};w) = - \sum_{i=1} ^2 y_i \log \sigma_i, \ \ \sigma_i : = \frac{e^{\hat{y}_i}}{e^{\hat{y}_1} + e^{\hat{y}_2}}, 
\end{equation}
where $y_i, i = 1, 2$ are the ground truth labels to indicate whether the image patch contains event of interest or not. 

In the following experiment section, we will see that labelling criteria and model architecture used in the classification head play an important role in the performance of the cell event recognition task.

\section{Experiments}\label{sec:experiments} 
We conduct experiments to show the benefit of using dense features from TAP pretraining and investigate empirically factors that impact the performance of the cell event recognition model. 

\paragraph{Data pre-processing}

The data we used in this paper are movie collected from live cells microscopy in our lab. The movie consists of $96$ 8-bit frames taken by the microscope with $15$ minutes gap between consecutive frames. All the data used in our analysis are taken from the same location in the sample. 

We label our cell microscopic images by creating a mask with pixels numbered according the event type. For example, in the area where nothing of interest happens, each pixel is labelled $0$. If a cell is diving or dying, each pixel in the area it covers will be labelled $1$. The labelling took around 8 person-hours. 

\begin{figure}[!h]
		\centering
		%
		%
		\begin{subfigure}[h]{0.85\textwidth}
			\includegraphics[width=\linewidth]{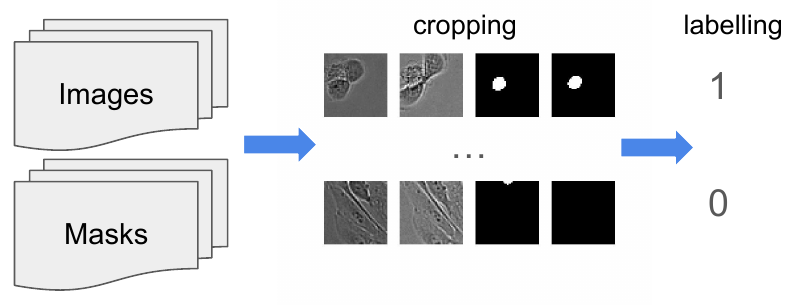}
		\end{subfigure}
		\vspace{1mm}
		\caption{Illustration of our data pre-processing. Starting with images and masks that are indexed by time, we take crops and get (crops, masks) pairs where crops are taken at the same location of the images at consecutive time points. We then apply our labelling criteria to turn the masks into binary labels.}
		\label{fig:data_preprocess_chart}
\end{figure}

To create our cell event type datasets from the images and their masks, we take random crops with size 48*48 from pairs of images that are consecutive in time. Then we create a binary label for each pair of masks based on the criterion that at least one of the mask should contain an event of interest. This is the default criterion for all the experiments below unless specified otherwise. Also, the location of the labelled pixels should be away from the edge of the crop. We approximate this by setting a threshold for the area of the labelled pixels and the area of labelled pixels in the mask needs to be above this threshold to quality as a positive event label. We illustrate our data pre-processing in a flow chart \ref{fig:data_preprocess_chart}. 
We randomly shuffle the crop pairs and split the them into training, validation and test sets with the ratio $0.6$, $0.2$ and $0.2$. We subsample the training set so that we have equal number of crop pairs with label $1$ and $0$. Table \ref{tbl:stats_dataset} shows a summary of our dataset.
\begin{table*}[h]
    \centering
        \scalebox{0.9}{
        \begin{tabular}{lcc}
        \toprule
        total crop pairs & 95000  \\ 
        training set (positive samples) & 14934 (7467)\\
        test set (positive samples) & 19000 (2470) \\
        validation set (positive samples) & 19000 (2458) \\ 
        Annotation area threshold & 40 \\
        crop size & 48*48 \\
         \bottomrule
        \end{tabular}
        }
    \caption{Summary of our dataset.}\label{tbl:stats_dataset}
\end{table*}
\paragraph{Pretraining using TAP}
We pretrain the feature map $f$ for TAP on all the frames from the movie. We provide details of the hyper parameters used and performance metrics in Section \ref{sec:details_TAP_pretraining} in the Appendix. Here we provide visualisations to show that the pretrained feature map has learned to pay attention to certain cell morphology such as divisions that is highly time sensitive. These visualisations are created using an attribution method Grad-CAM \citep{SelvarajuGrad-CAM2017} which in our case highlights most important regions in the input image for predicting time arrow. In the next section, we will provide comprehensive comparisons to show that TAP features are useful for cell event recognition.  
\begin{figure*}[!h]
        \centering
        %
        %
        \begin{subfigure}[h]{0.49
        \textwidth}
    \includegraphics[width=\linewidth]{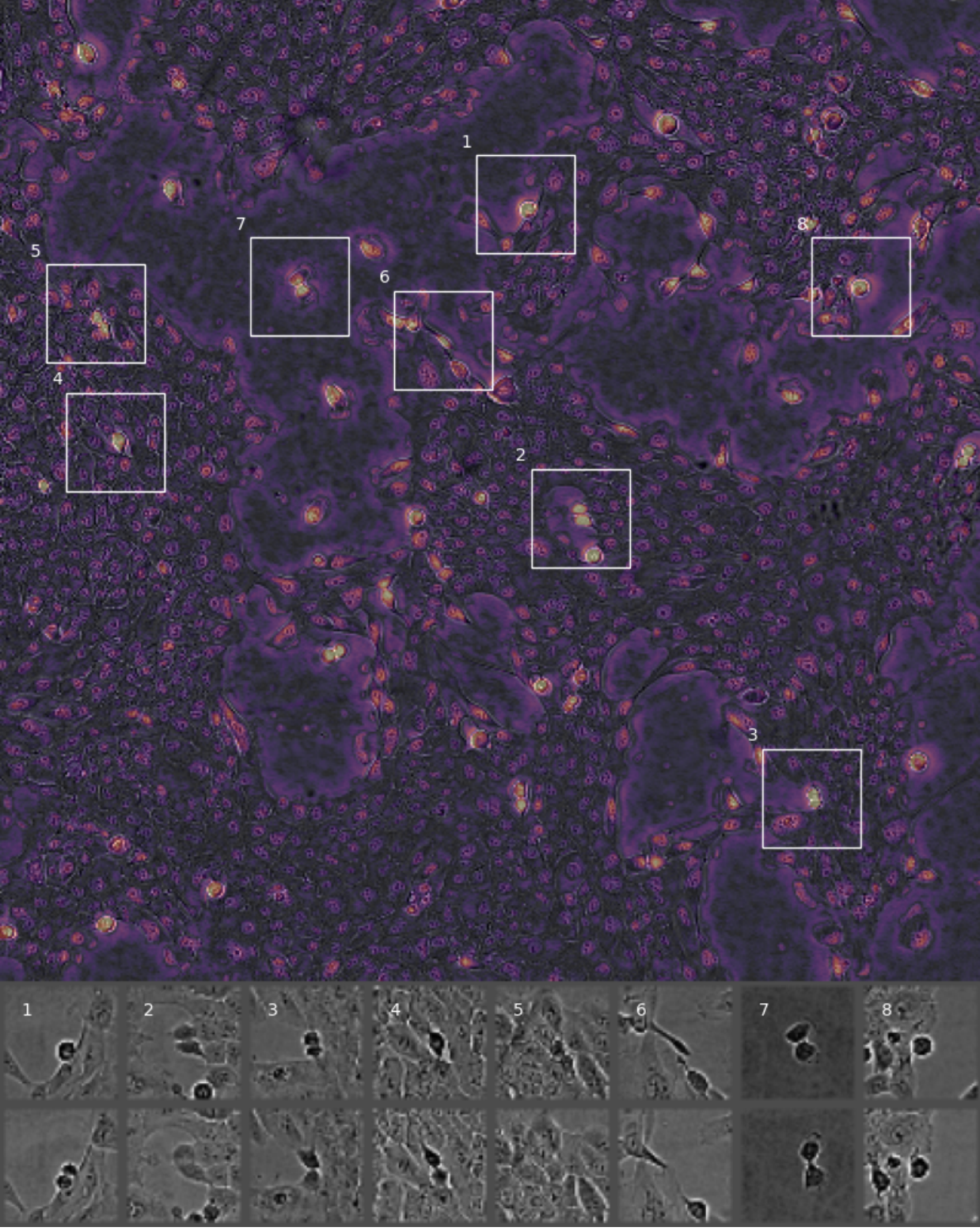}\\
            \caption{}\label{fig:grad cam training examples}
        \end{subfigure}
        \begin{subfigure}[h]{0.49\textwidth}
            \includegraphics[width=\linewidth]{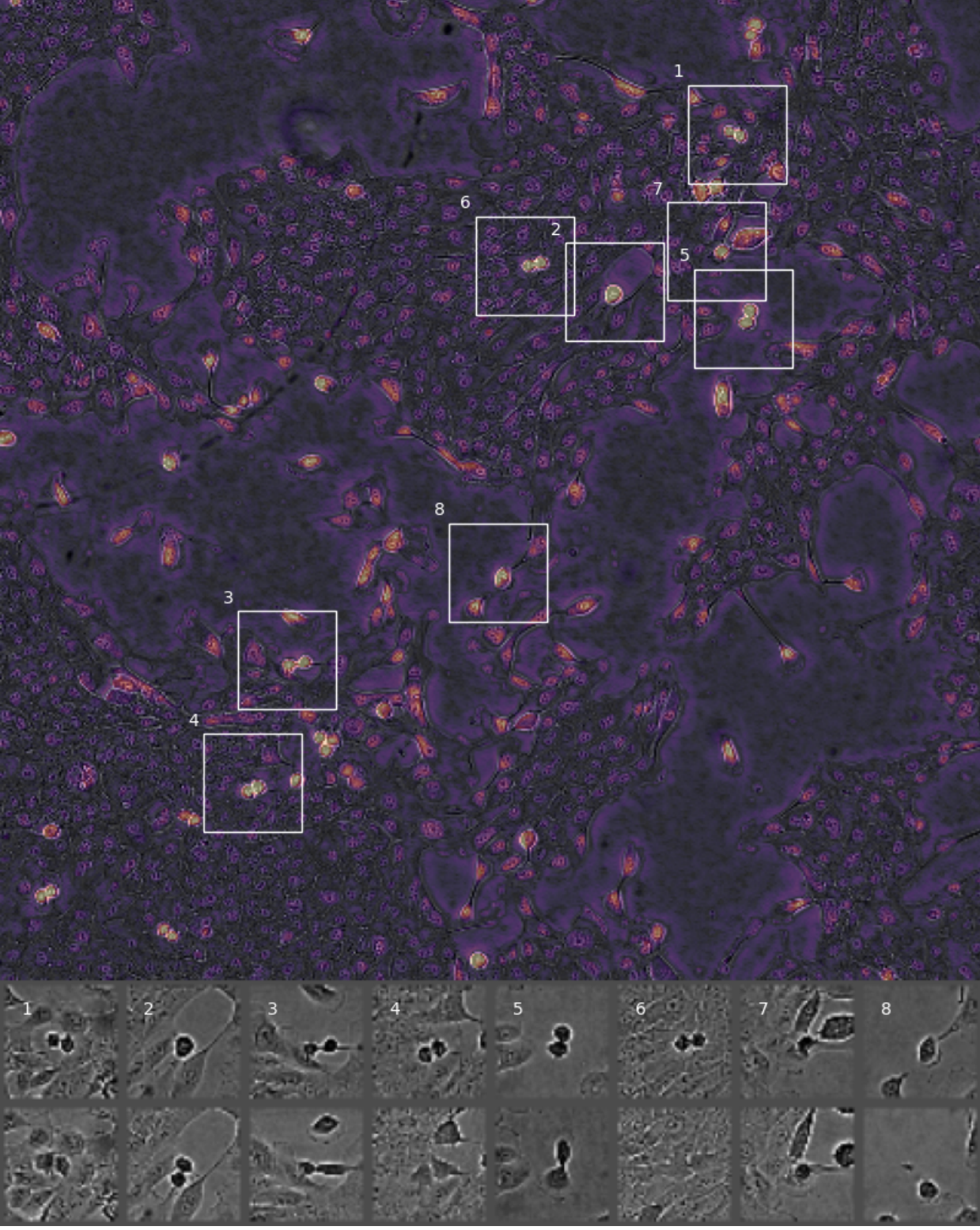}\\
            \caption{}\label{fig:grad cam validation examples}
        \end{subfigure}
        \caption{Example images overlaid with attribution map from Grad-CAM \citep{SelvarajuGrad-CAM2017}. Figure \ref{fig:grad cam training examples} shows one image from the training set and Figure \ref{fig:grad cam validation examples} shows one from the validation set which is taken from the same experiment but a different location of the live cell sample. The numbered crops are showing the top $8$ regions ranked by the Grad-CAM scores, so the higher the more influential the pixels in the region have on the time arrow prediction. The bottom rows show the view at current time point and one frame afterwards. We can see many of these highlighted regions contain changes in cell morphology resulted from divisions.}
        \label{fig:grad_cam visualisations}
\end{figure*}
We also observe through experiments that the validation accuracy of time arrow prediction is influenced by the size of `content view' in a crop, i.e. the amount of visual information the model sees in the crop. The larger the content view, the more accurate the prediction is. In our setting, the size of the `content view' is affected by the crop size (measured by pixel dimensions) and the resolution of the image. When we fix the resolution of the image, taking a larger crop will give us a larger `content view'. When we reduce the resolution of the image, a crop of the same size will give us a larger `content view'. However, we also notice that reducing the resolution of the image can reduce the accuracy of the prediction. We provide some experimental results on how the accuracy of TAP can be affected different crop sizes and resolutions in Section \ref{supp sec:tap accuracy} in the supplementary material.

\paragraph{Do TAP features actually help?}
Besides supporting observations from the previous section, to find out whether the dense features learned from TAP using U-net can really help the downstream task of cell event recognition, we consider the following training strategies and compare their results:
\begin{enumerate}
    \item [A0] Train a model using a linear classification head and the pre-trained dense representations from TAP as described in Section \ref{sec:cell_event_recog}. 
    \item[B0] Initialise the U-net model randomly and use the representation from this untrained model (fixed) when we train the event classification head. 
    \item[C0] Similar to [B0] but the entire combined model gets trained from end-to-end for the classification task without fixing the randomly initialised weights for the feature map. 
\end{enumerate}
We also repeat A0 - C0 with the linear head replaced by a non-linear (ResNet) head and we name the results A1 - C1. Regarding the ResNet head, we will only be considering a simple residual network with one convolutional layer followed by one residual block.  

From the results summarised in Table 
\ref{tbl:TAP_features_comparisons}, we see that when we use the TAP features and train a ResNet head on top of those features we get the best performance measured by precisions and recalls for class 0 and 1. We also observe that for a fixed feature learning strategy (e.g. using TAP feature or randomly initialised weights, fixed or trainable), we get better performance by using the ResNet head compared to the linear head, demonstrating the benefit of using a non-linear classification head. 
\begin{table*}[!h]
		\centering
		\caption{Results from comparing different training strategies with or without TAP features. Using TAP features with ResNet as classification head performs best in all confusion matrix entries (i.e. precision and recall for class 0 (no event) and class 1 (with event). For each group, we report the mean and standard deviation of the precision and recall for $10$ runs. When we initialise the U-net randomly, we use Kaiming uniform initialisation \citep{HeDelvingDeep2015}. We fine tune the linear head for 10 epochs and ResNet head for 30 epochs.}\label{tbl:TAP_features_comparisons}
		\vspace{8pt}
		\scalebox{0.8}{
		\begin{tabular}{lcccccc}
		\toprule
		    group & feature & cls head & prec 0 & rec 0 & prec 1 & rec 1\\
		     \midrule
	          A0 & TAP & linear & $0.98 \pm 0.0$ & $0.87 \pm 0.03$ & $0.5 \pm 0.04$ & $0.85 \pm 0.03$\\ 
                A1 & TAP & Resnet & $\textbf{0.98} \pm \textbf{0.0}$ & $\textbf{0.94} \pm \textbf{0.01}$ & $\textbf{0.7} \pm \textbf{0.03}$ & $\textbf{0.89} \pm \textbf{0.03}$ \\ 
                B0 & random (fixed) & linear & $0.95 \pm 0.04$ & $0.6 \pm 0.32$ & $0.34 \pm 0.2$ & $0.67 \pm 0.36$   \\ 
                B1 & random (fixed) & Resnet & $0.97 \pm 0.01$ & $0.87 \pm 0.07$ & $0.5 \pm 0.08$ & $0.8 \pm 0.05$ \\
                C0 & random (trained) & linear & $0.97 \pm 0.0$ & $0.92 \pm 0.01$ & $0.6 \pm 0.02$ & $0.82 \pm 0.01$ \\
                C1 & random (trained) & Resnet & $0.98 \pm 0.01$ & $0.9 \pm 0.04$ & $0.57 \pm 0.08$ & $0.86 \pm 0.05$ \\
		     \bottomrule
		\end{tabular}
		} 
\end{table*}

\begin{figure}[!h]
		\centering
		%
		%
		\begin{subfigure}[h]{0.95\textwidth}
			\includegraphics[width=\linewidth]{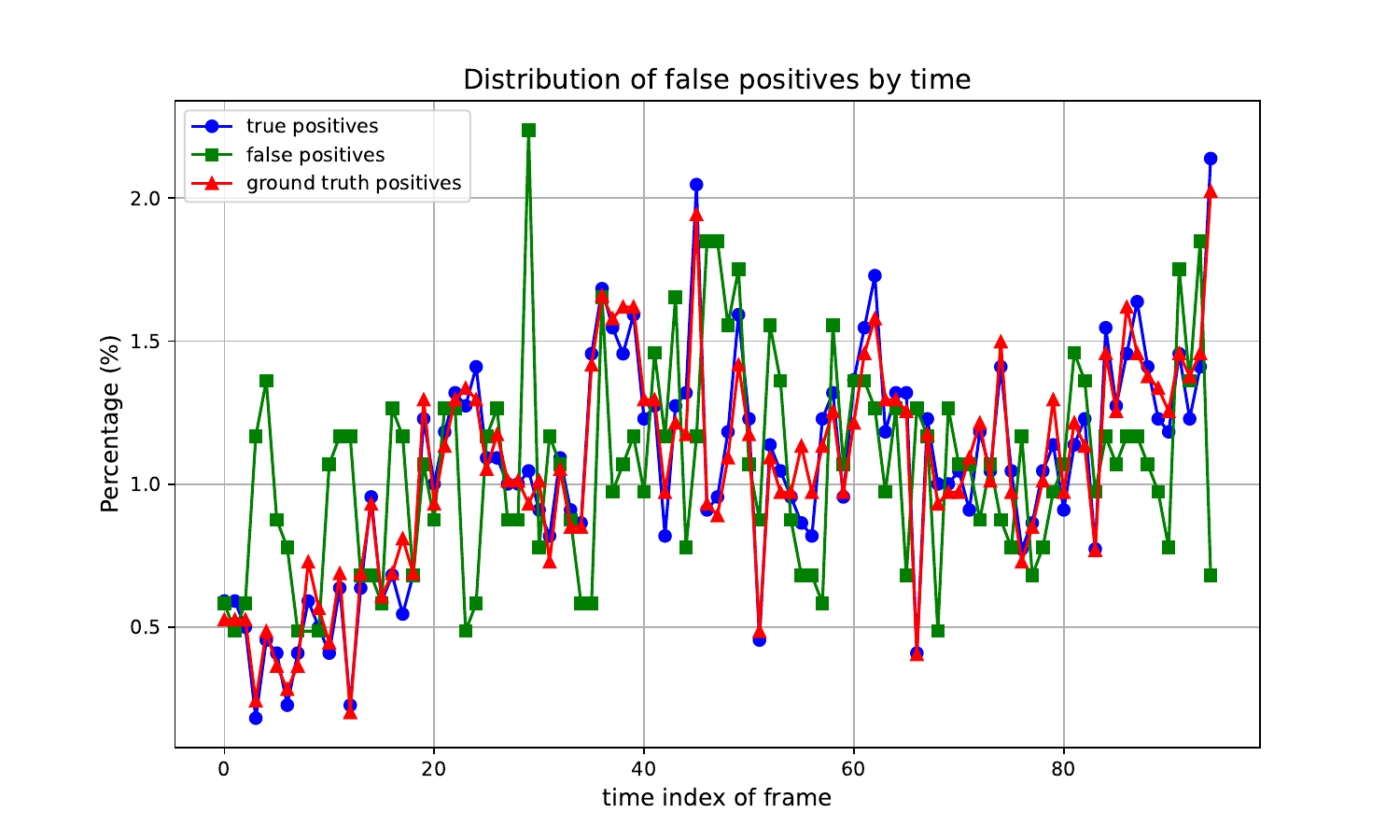}
		\end{subfigure}
		\vspace{1mm}
		\caption{Distribution of false positive predictions by time. The percentage at a specific time index is calculated by dividing the number of false positive predictions at that time point by the total number of false positives made by the model over the entire test dataset. We also plot the true positives and ground truth positives for comparison.}
		\label{fig:false_positives_by_time}
\end{figure}

\begin{figure}[!h]
		\centering
		%
		%
		\begin{subfigure}[h]{0.95\textwidth}
			\includegraphics[width=\linewidth]{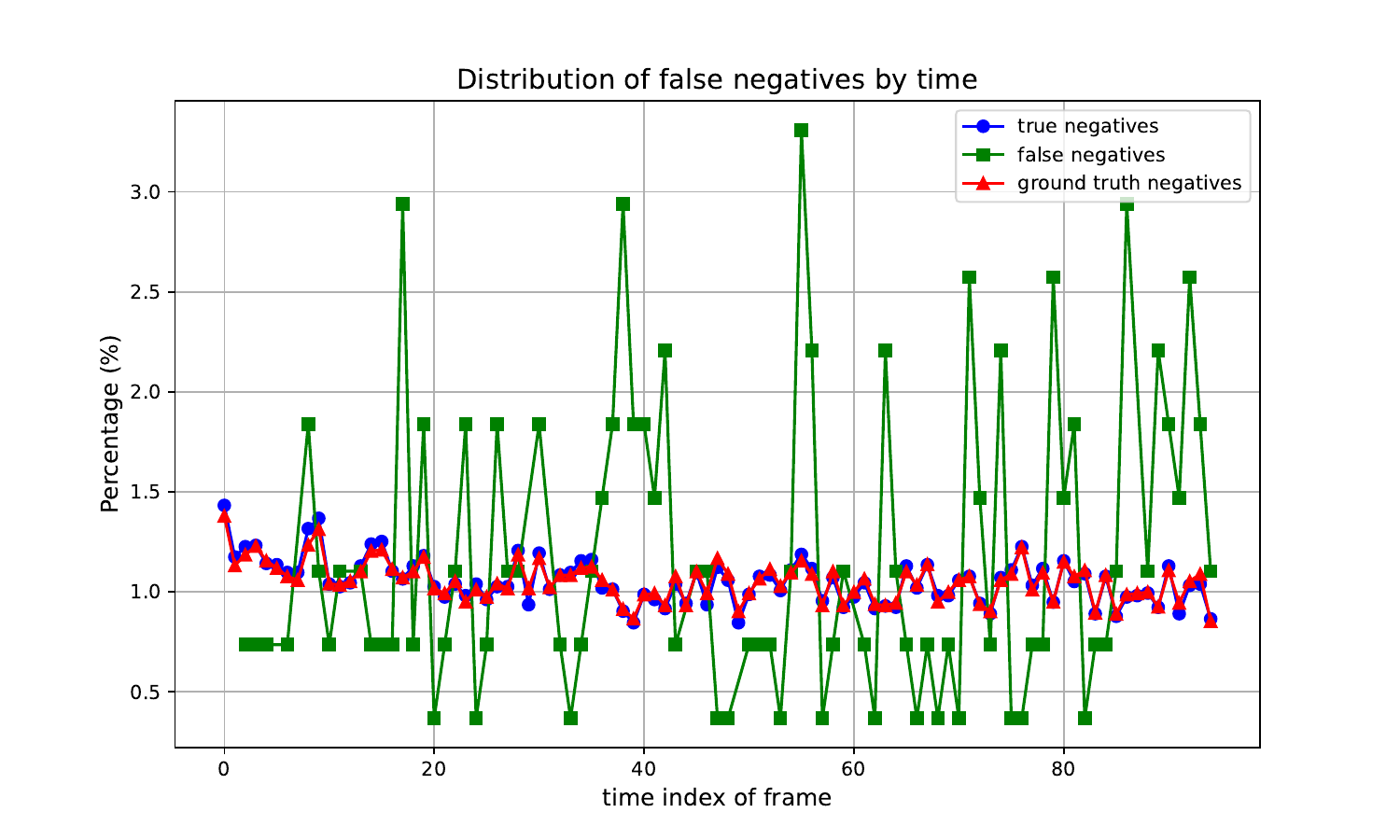}
		\end{subfigure}
		\vspace{1mm}
		\caption{Distribution of false negative predictions by time. The percentage at a specific time index is calculated similarly over the negative datapoints. We also plot the true negatives and ground truth negatives for comparison.}
		\label{fig:false_negatives_by_time}
\end{figure}

\paragraph{Examining the true positives and the mistaken predictions}
We provide analysis on the true positives, false positives and false negatives from a model trained using TAP features and fine tuned using ResNet head (i.e. strategy using in group A1 in 
\ref{tbl:TAP_features_comparisons}). We show the overall performance measured by the confusion matrix, precision and recall scores in Table \ref{tbl:confusion_matrix}. Furthermore, we have also considered temporal patterns of the mistaken predictions which we give more details in Figure \ref{fig:false_positives_by_time} and  \ref{fig:false_negatives_by_time}. 
\begin{table*}[h]
    \centering
    \begin{minipage}[b]{0.45\linewidth} 
        \centering
        \vspace{8pt}
        \scalebox{0.9}{
        \begin{tabular}{lccc}
        \toprule
        & predicted 0 & predicted 1 \\ 
        \midrule
        actual 0 & 15502 & 1028 \\
        actual 1 & 272 & 2198 \\
         \bottomrule
        \end{tabular}
        }
    \end{minipage}
    \hspace{0.05\linewidth} 
    \begin{minipage}[b]{0.45\linewidth} 
        \centering
        \vspace{8pt}
        \scalebox{0.9}{
        \begin{tabular}{lccc}
        \toprule
        & precision & recall \\ 
        \midrule
        class 0 & 0.98 & 0.94 \\
        class 1 & 0.68 & 0.89 \\
         \bottomrule
        \end{tabular}
        }
    \end{minipage}
    \caption{Confusion matrix for a model trained from using fixed TAP featrures and fine-tuned with ResNet head (i.e. strategy in group A1 in Table \ref{tbl:TAP_features_comparisons}) on the test dataset.}\label{tbl:confusion_matrix}
\end{table*}
\begin{enumerate}
    \item From Figure \ref{fig:false_positives_by_time}, we observe that the distribution of true positives follows closely with the actual (i.e. ground truth) positives, showing the high level accuracy of the model predictions. At the start of the image sequence (e.g. in between frame $0$ and $20$), false positives surpass true positives (and actual positives) by 
   more compared to later frames where they follow the same trend more closely. We think that this is attributed to the fact that the number of cells and events increase over time.

   \item From Figure \ref{fig:false_negatives_by_time}, we observe that the distribution of ground truth negatives is relatively uniform in time, compared to ground truth positives. Again, the true negatives follow closely with the ground truth negatives, which shows the accuracy of our model on predicting non events. Notice that the percentage of the false positives stay uniformly low for most of the frames (recall that false positives are the difference between ground truth positives and true positives, which we see are quite close in Figure \ref{fig:false_positives_by_time}). 
    
\end{enumerate}
\paragraph{Criteria for events labelling }
What is the criterion for labelling an image pair to contain an event of interest that is most suitable for our imaging dataset and characterises the cell behaviour well? We believe there is no clear-cut answer to this question. In Table \ref{tbl:illustration_labelling_criteria}, we provide four examples to illustrate how the same examples might be labelled differently under the labelling criteria we considered. 
\begin{enumerate}
    \item [A] No threshold for the labelled pixel area. At least one of crops in the pair must contain events of interest. 
    \item [B] Similar to A but setting a positive threshold.
\end{enumerate}
\begin{table}[!h]
    \centering
    \scalebox{0.7}{
    \begin{tabular}{cccc}
    \toprule
     & Any size, both & any size, either & size filter, either \\ 
    \midrule
    \adjustbox{valign=m}{\includegraphics[scale=0.3]{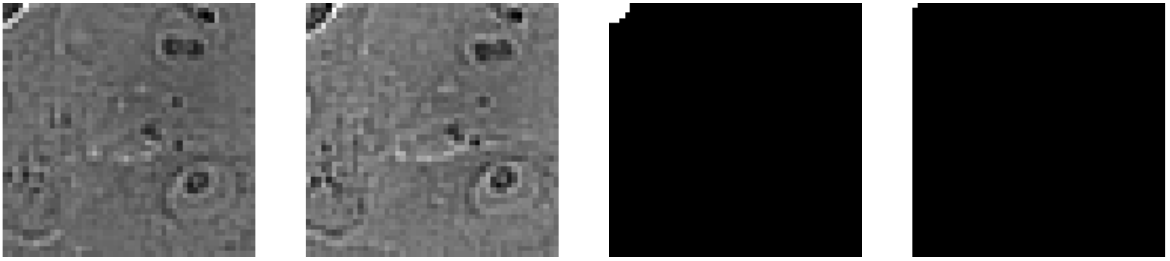}} & 1 & 1 & 0 \\ 
    \adjustbox{valign=m}{\includegraphics[scale=0.3]{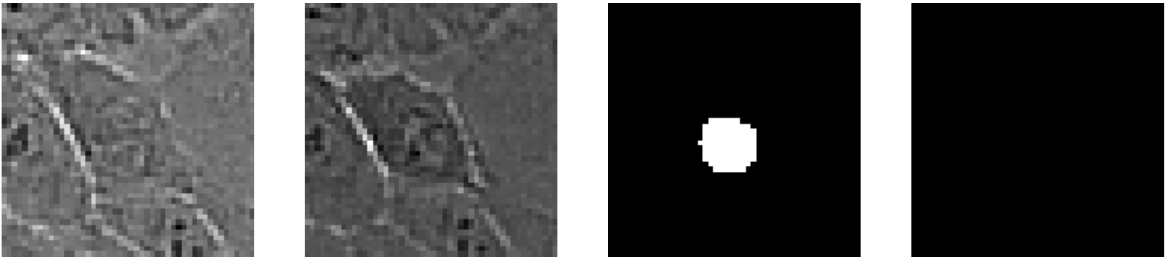}} & 0 & 1 & 1 \\ 
    \adjustbox{valign=m}{\includegraphics[scale=0.3]{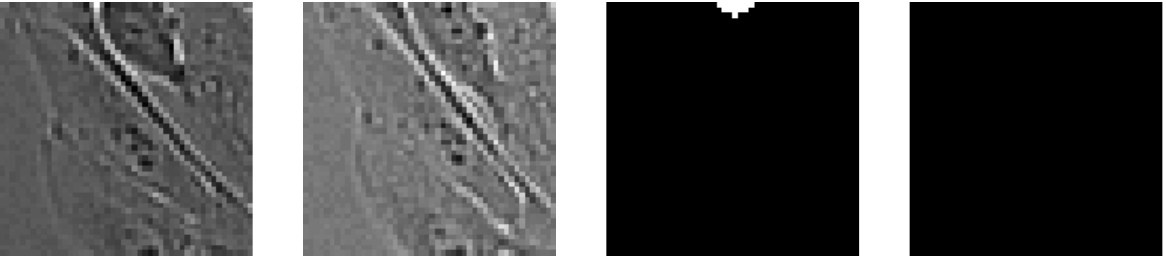}} & 0 & 1 & 0 \\ 
    \adjustbox{valign=m}{\includegraphics[scale=0.3]{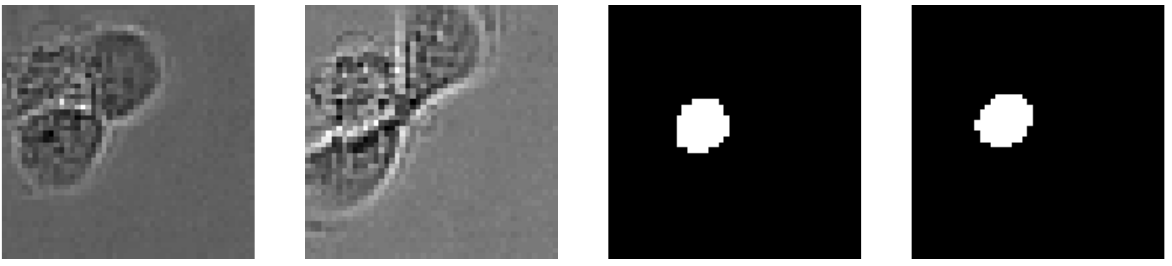}} & 1 & 1 & 1 \\ 
    \bottomrule
    \end{tabular}
    }
    \caption{Illustrations of how different labelling criteria can result in the same examples being labelled differently. In the first row, the annotated part appear at the edge in both crops but do not pass the size threshold. In the second row, only the first crop contains annotation that is located away from the edge. In the third row, only the first crop contains annotation but it is located at the edge and does not pass the size threshold. In the fourth row, both crops contain annotations which are located centrally.}
    \label{tbl:illustration_labelling_criteria}
\end{table}

Now we compare the confusion matrices of the model prediction according to different criteria for event labelling. We consider the following variants of criteria and train and evaluate models with either linear or ResNet head with data labelled according these criteria.The results are shown in Table \ref{tbl:labelling_criteria_comparisons}. We observe that when moving from `any size, both` to `any size, either', we are effectively lowering the bar for a sample (i.e. a pair of image crops) to be given a positive label. This results in more positive samples in the training data and leads to improved test set performance for our model regardless of the classification head used. We further impose a size filter so that we only label a crop pair positive when the event occurs away from the boundary and is close to the centre. This `size filter, either' criteria has reduced the number of positive samples but has further increased the test set performance of the model using either linear or ResNet head. We believe applying the size filter has increased the quality of the training samples by reducing the ambiguity of the label which makes it easier for the model to distinguish positive samples from negative ones. We also notice that for the same labelling criteria, model with ResNet head always performs better than that with linear head, demonstrating the benefit of increasing the capacity of the classification head. 
\begin{table*}[h]
		\centering
		\caption{Performance of models with either linear or ResNet heads under different labelling criteria for positive samples. `Any size' refers to that criterion that allows any size of labelled area in the mask, while `size filter' refers to applying a size threshold of $40$. `both' and `either' refer to whether to label a pair of crops when either one or both of them contain a positive label. We provide the number of positive samples in training and testing data in `positive events'. }\label{tbl:labelling_criteria_comparisons}
		\vspace{8pt}
	  \scalebox{0.75}{
		\begin{tabular}{lcccccc}
		\toprule
		    labelling criteria & positive samples & cls head & prec 0 & rec 0 & prec 1 & rec 1\\
		     \midrule
	          any size, both  &       5858, 1867           &linear & $0.98 \pm 0.0$ & $0.85 \pm 0.01$ & $0.39 \pm 0.02$ & $0.87 \pm 0.01$\\ 
              any size, both  &   	  5858, 1867		   & Resnet & $\textbf{0.99} \pm \textbf{0.0}$  & $0.91 \pm 0.04$ & $0.52 \pm 0.07$  & $\textbf{0.89} \pm \textbf{0.03}$ \\ 
              any size, either & 	  	9946, 3284			&linear & $0.96 \pm 0.0$ & $0.8 \pm 0.03$ & $0.47 \pm 0.04$ & $0.83 \pm 0.03$   \\ 
              any size, either & 		9946, 3284			&Resnet & $0.97 \pm 0.0$ & $0.93 \pm 0.02$ & $\textbf{0.72} \pm \textbf{0.04}$ & $0.86 \pm 0.02$ \\
              size filter, either & 	7467, 2470		    &linear & $0.98 \pm 0.0$ & $0.87 \pm 0.03$ & $0.5 \pm 0.04$ & $0.85 \pm 0.03$ \\
              size filter, either & 	7467, 2470			 &Resnet & $0.98 \pm 0.0$ & $\textbf{0.94} \pm \textbf{0.01}$ & $0.7 \pm 0.03$ & $\textbf{0.89} \pm \textbf{0.03}$ \\
		     \bottomrule
		\end{tabular}
		} 
\end{table*}

\paragraph{Model calibration}
It is shown in \cite{Guo2017OnCalibration} that deep neural networks can get over confident at predictions even when they are incorrect, due to mis-calibration. Confidence calibration is the process of predicting the probability estimates that represent the correctness likelihood of the classification \citep{Guo2017OnCalibration}. We believe confidence calibration is an important post-processing step after the mode has been trained that can benefit applications such as assisting lab scientists in human-in-the-loop cell behaviour annotations and analysis. In this experiment, we apply a calibration method called temperature scaling from \cite{Guo2017OnCalibration} and show reliability diagrams (i.e. diagrams showing average accuracy at each confidence level) before and after. Temperature scaling is the process of learning a scaler that is multiplied to the confidence score (Sigmoid of the logits in the case of binary classification) so that the expected calibration error (ECE) is minimised.  
\begin{figure*}[!ht]
		\centering
		%
		%
		\begin{subfigure}[h]{0.49\textwidth}
		    \begin{minipage}[c]{0.85\linewidth}
		    \includegraphics[width=1.1\linewidth]{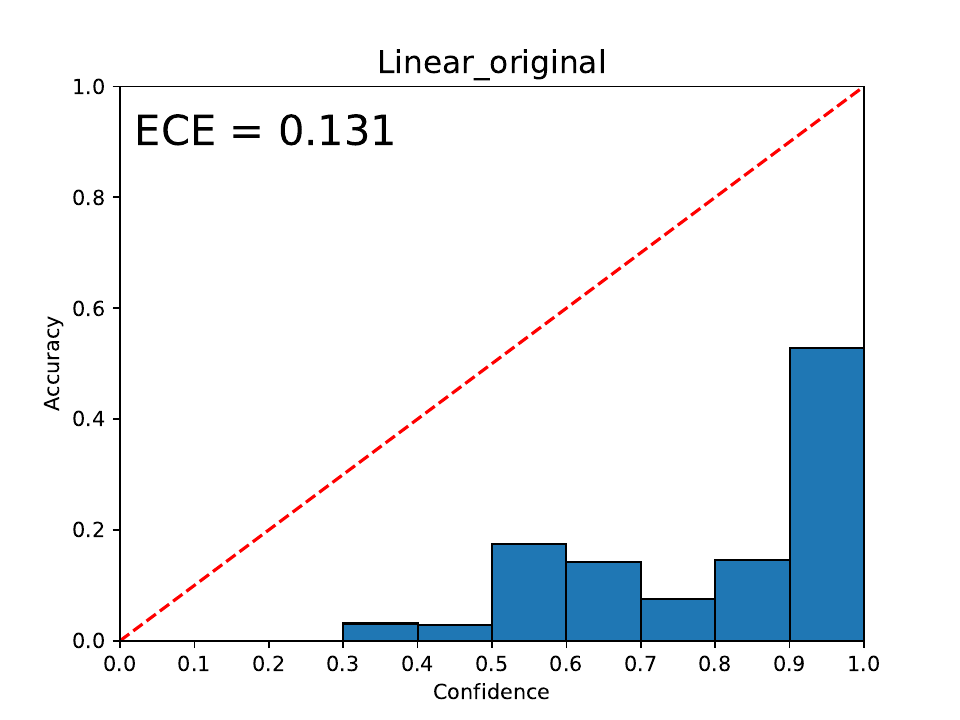}\\
		    \end{minipage}\caption{}\label{fig:calibration linear original}
		\end{subfigure}\hfill
		\begin{subfigure}[h]{0.49\textwidth}
		    \begin{minipage}[c]{0.85\linewidth}
		    \includegraphics[width=1.1\linewidth]{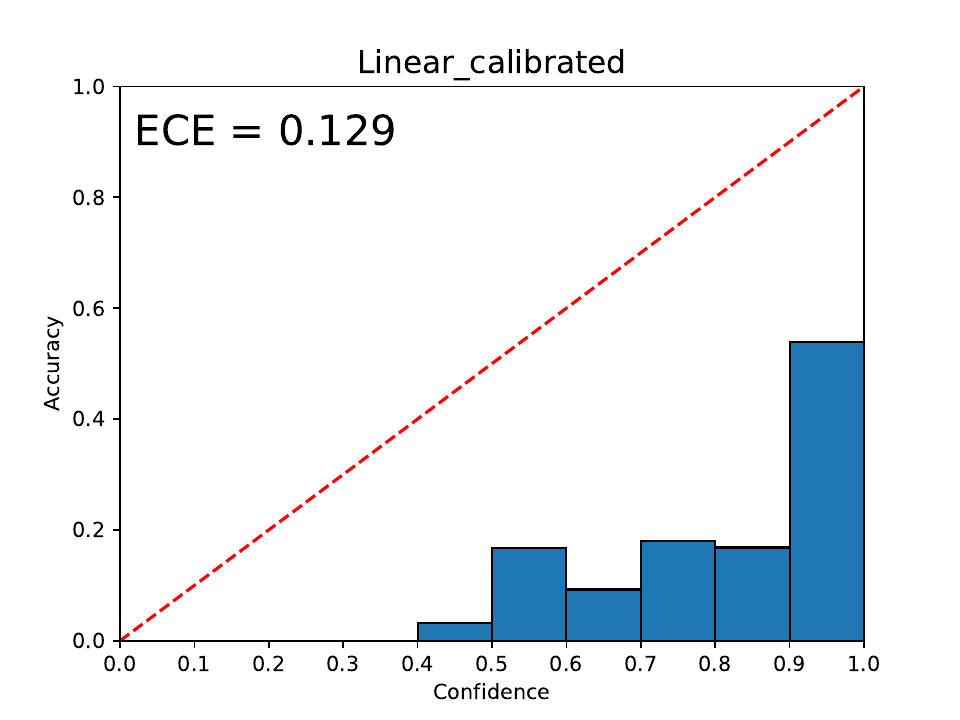}\\
		    \end{minipage}\caption{}\label{fig:calibration linear calibrated}
		\end{subfigure}\hfill
		\begin{subfigure}[h]{0.49\textwidth}		
		    \begin{minipage}[c]{0.85\linewidth}
		    \includegraphics[width=1.1\linewidth]{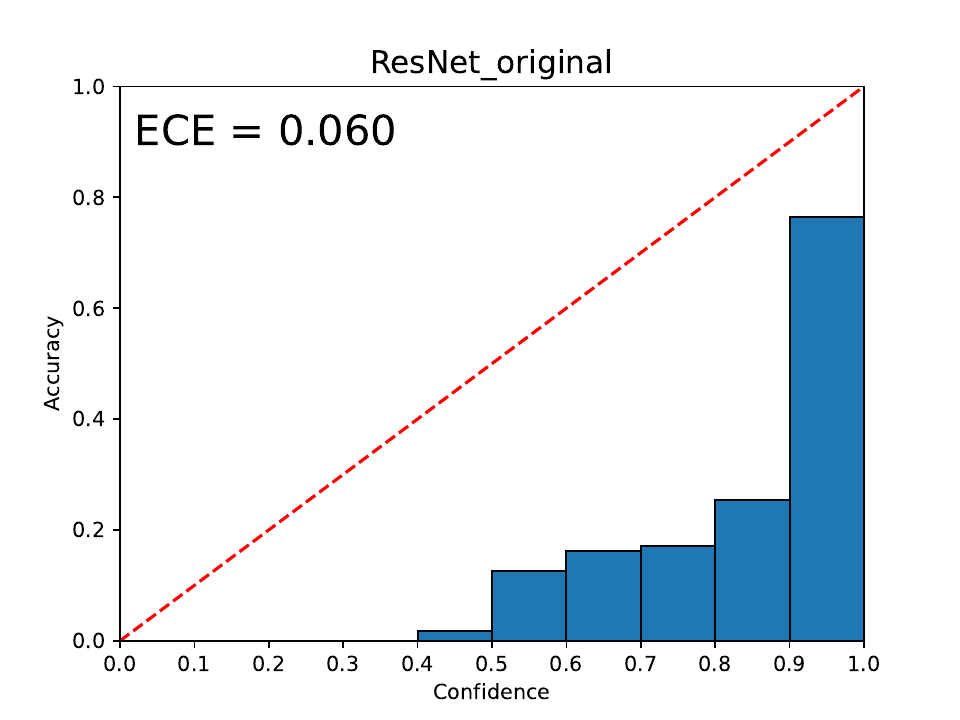}\\
		    \end{minipage}\caption{}\label{fig:calibration resnet original}
		\end{subfigure} \hfill
        \begin{subfigure}[h]{0.49\textwidth}
		
		    \begin{minipage}[c]{0.85\linewidth}
		    \includegraphics[width=1.1\linewidth]{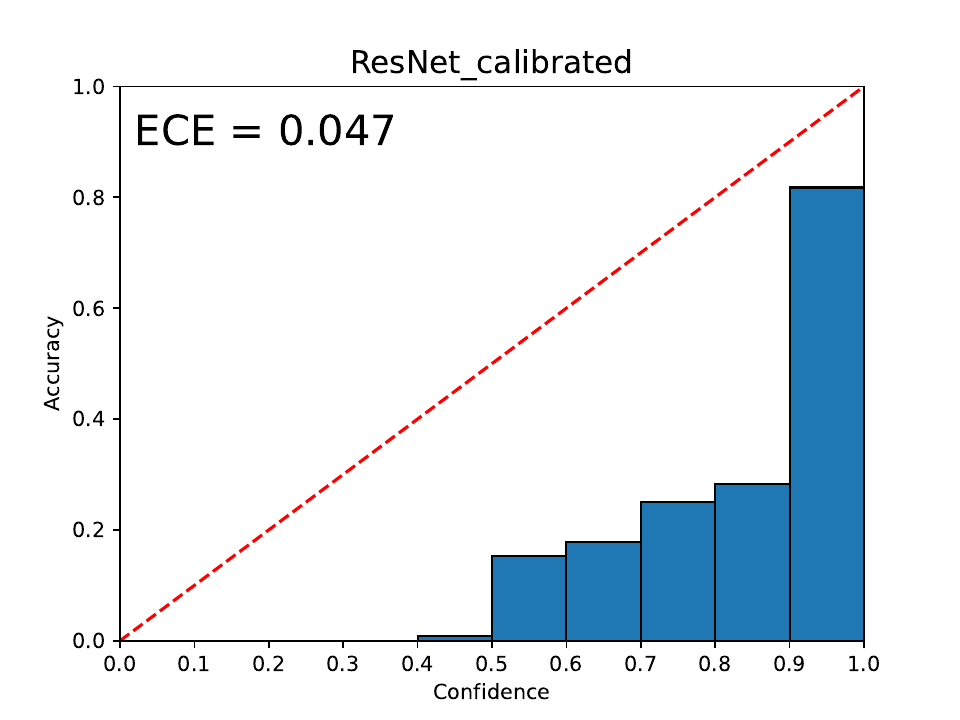}\\
		    \end{minipage}\caption{}\label{fig:calibration resnet calibrated}
		\end{subfigure} 
		\vspace{1mm}
		\caption{Reliability diagrams of the original and calibrated models. The red dashed line indicates perfect calibration where the accuracy (for class 1) is equal to the confidence. At a fixed confidence level, the prediction is over-confident if the accuracy is below the dashed line. In the left column, \ref{fig:calibration linear original} and \ref{fig:calibration resnet original} show the reliability diagrams for the model with linear and ResNet head before calibration using temperature scaling, and the right column show those after calibration. The expected calibration error (ECE) is shown in the top left corner.}
		\label{fig:reliability diagrams}
\end{figure*}
We plot reliability diagrams in Figure \ref{fig:reliability diagrams} with the ECE scores. 

We observe that by using temperature scaling, we are able to reduce ECE and adjust the confidence of predictions to be closer to the diagonal for the model with either the linear head (Figure \ref{fig:calibration linear original}, \ref{fig:calibration linear calibrated}) or the ResNet head (Figure \ref{fig:calibration resnet original}, \ref{fig:calibration resnet calibrated}). Also observe that the model with the ResNet head is better calibrated at the start compared to the model with the linear head.  

\section{Related work}\label{sec:related_work}
\paragraph{Segmentation based approach using supervised learning}
One direction to approach the cell event recognition in live-cell microscopy is to apply deep-learning based detection and segmentation methods that fully supervised. The segmentation methods can be classified into two basic categories: semantic segmentation aims to classify each pixel into a semantic class (e.g. nucleus or background) whereas instance segmentation tries to create a label mask for each individual cell. Examples of works following this approach include \cite{GreenwaldWholeCell2022}, \cite{StringerCellpose2020}, \cite{Cicek3DUnet2016}, \cite{WeigertCellDetectionwithStar2018} and \cite{WeigertStarConvex2020}. Despite the progress made in the direction, these methods usually require large amount of high-quality annotated data, a challenge magnified by the task of live-cell event recognition because of the extra time dimension.   

\paragraph{Time arrow prediction and self-supervised representation learning}
The work \cite{PickupSeeingtheArrow2014} proposed methods to detect time arrow in videos (i.e. whether a video is running forward or backward in time). They demonstrated their methods on a selection of YouTube video clips, and on natively-captured sequences. Although it does not use deep neural network for learning feature maps, the idea has been applied to various computer vision problems such as action recognition \citep{WeiLearningandUsingtheArrow2018}, \citep{BroomeReccurAttend2023} and video anomaly detection \citep{WangVideoAnomalyDetection2022}.  

Self-supervised representation learning (SSRL) methods aim to provide powerful, deep feature learning without the requirement of large annotated data sets. The choice of pretext task in SSRL determines the invariance of the learned representation (or feature maps) and thus how effective it is for different downstream tasks \citep{EricssonSelfSupervisedRepresentation2022}. \cite{GallusserSelfsupervisedDense2023} considers time arrow prediction as the pretext task and use the learned representation for downstream tasks in live-cell microscopy including detection and segmentation of dividing cells and cell state classification.

We follow the approach in \cite{GallusserSelfsupervisedDense2023} and carry out extensive experiments to demonstrate the benefits using TAP as the feature maps for cell event recognition and highlight a few issues that we believe can serve as guidelines for applying SSRL using TAP in live-cell microscopy. 

\section{Limitations and future work}\label{sec:limit_future_work}
When we examined the false positive predictions when we use the TAP features with the linear class head, we found that polarised cell translocation, especially a high-contrast cell at the edge of a crop, change of contrast (for example dead bodies of cells), transient debris or artefact may contribute to false positive predictions. We think these limitations can be attributed to the use of TAP features which might be sensitive to all signals indicating time arrow and not necessarily relevant to cell states. In our comparison experiment summarised in Table \ref{tbl:TAP_features_comparisons}, we have shown that false positives can be reduced by switching from linear head to non-linear head such as ResNet (where recall for class 0 increases from group A0 to A1).

It is reasonable to believe that having a time-invariant prediction head will increase the performance of the model with the same number of training epochs, because the task of cell event recognition itself is time-invariant. The event of death or division should not change if we play the movie forward or backward. However, we notice in our experiments that making the prediction head to be time-invariant does not improve the performance of the model on the event recognition task. It remains interesting to find out if the alignment between time-equivariance and time-invariance of the pretext task and of the downstream task is necessary.

We haven't investigated other pretext task besides TAP. It would be interesting to consider representations obtained by generative models such as diffusion models (\cite{Sohl-DicksteinDeepUnsupervisedLearning2015}, \cite{SongGenerativeMB2019}, 
\cite{HoDenoisingDiffusion2020}). 
In addition, we are interested in potential performance gain by replacing the U-net by vision transformers \citep{KolesnikovImageisWorth2021} as the backbone in learning TAP features.

\section{Conclusion}
We demonstrate through extensive experiments and analysis that time arrow prediction is an effective pretext task to learn representations in self-supervised way that can greatly benefit the downstream task of cell event recognition in live-cell microscopy where the temporal dimension creates additional challenges. 

The code and dataset used in this work will be made publicly available. 
\bibliographystyle{plainnat}
\bibliography{ref}

\newpage
\appendix{\textbf{Supplementary Material to `Self-supervised Representation Learning for Cell Event Recognition through Time Arrow Prediction'}}

\section{Loss function for the time arrow prediction}
For reference, we summarise the construction of the loss function for TAP based on \cite{GallusserSelfsupervisedDense2023}. For image patches $(x_t, x_{t+\Delta t}$ from frames at $t$ and $t + \Delta t$, we flip the order of each pair with probability $0.5$ and assign a label $y$ according to its order in time. Then we compute the dense features $z_t := f(x_t), z_{t + \Delta t} := f(x_{t+\Delta t})$ from a fully convolutional feature map $f$ which is taken to be a three-block U-net in \cite{GallusserSelfsupervisedDense2023} and in this work. Then we produce logits $\hat{y} = u([z_t, z_{t + \Delta t}])$ using a time arrow prediction head $u$. Both $f$ and $u$ are trained jointly to minimise the loss 
\begin{equation}
    \CL(y, \hat{y}) + \lambda \CL_{Decorr}(z),
\end{equation}
where $\CL(y, \hat{y})$ is the cross entropy loss on the softmax values of $\hat{y}$ and $\CL_{Decorr}(z)$ is a loss term that penalises correlations across feature channels via maximising the diagonal of the softmax-normalised correlation matrix $a_{ij}$:
\begin{equation}
    \CL_{Decorr}(z) = - \frac{1}{C}\sum_{i=1} ^C \log a_{ii}, \ \  a_{ij}  = \frac{e^{z_i \cdot z_j / \tau}}{\sum_{j=1} ^C e^{z_i \cdot z_j / \tau}}. 
\end{equation}
Here $C$ is the number of channels, $\tau$ is temperature parameter which is fixed to be $0.2$, and $\lambda = 0.01$ in training and inference. Similar to \cite{GallusserSelfsupervisedDense2023}, We adopt the definition of the TAP head $u$ to be time equivariant, which means that for any $z_t, z_{t + \Delta t}$ such that $ u([z_t, z_{t + \Delta t}]) = [\hat{y_t}, \hat{y}_{t + \Delta t}]$, we have $ u([z_{t + \Delta t}, z_t]) = [\hat{y}_{t + \Delta t}, \hat{y}_t]$.

\section{Details of learning the feature map from TAP pretraining}\label{sec:details_TAP_pretraining}
\begin{figure*}[!h]
        \centering
        %
        %
        \begin{subfigure}[h]{0.49\textwidth}
            \begin{minipage}[c]{0.85\linewidth}
            \includegraphics[width=1.1\linewidth]{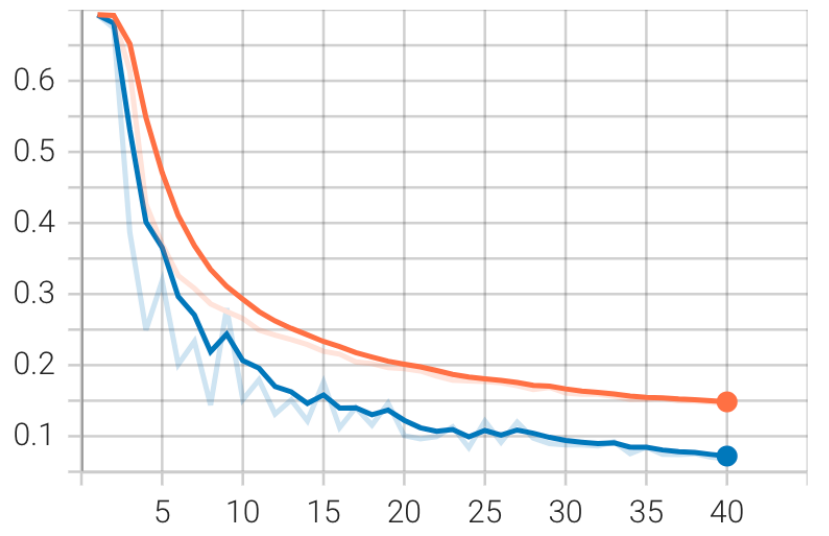}\\
            \end{minipage}\caption{}\label{fig:TAP pretraining training loss}
        \end{subfigure}\hfill
        \begin{subfigure}[h]{0.49\textwidth}
        
            \begin{minipage}[c]{0.85\linewidth}
            \includegraphics[width=1.1\linewidth]{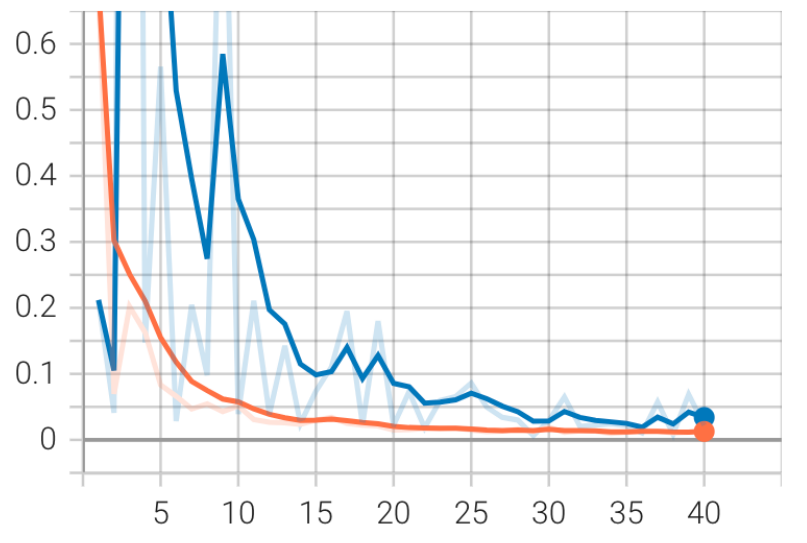}\\
            \end{minipage}\caption{}\label{fig:TAP pretraining training decorrelation loss}
        \end{subfigure}\hfill
        \begin{subfigure}[h]{0.49\textwidth}

            \begin{minipage}[c]{0.85\linewidth}
            \includegraphics[width=1.1\linewidth]{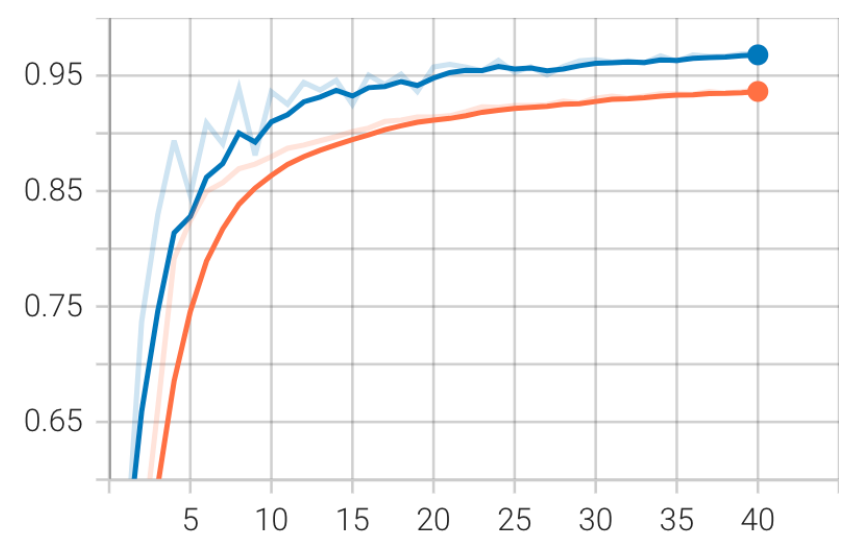}\\
            \end{minipage}\caption{}\label{fig:TAP pretraining accuracy}
        \end{subfigure} \hfill
        \vspace{1mm}
        \caption{Figures \ref{fig:TAP pretraining training loss}, \ref{fig:TAP pretraining training decorrelation loss} and \ref{fig:TAP pretraining accuracy} show total losses, decorrelation losses and accuracy during training. Orange curves refer to the corresponding metric on the training set and blue on the validation set.}
        \label{fig:TAP pretraining metrics}
\end{figure*}

In training, we set the patch size to be 96*96.  Due to limited GPU memories we have, we are not able to choose a bigger patch size. We train the U-net with a convolutional time-equivariant classification head for TAP on the live-cell movie data taken at one location of the sample and validate it on one from a nearby location. The training data consist of $96$ frames of 1024*1024 8-bit images and the validation data consist of $97$ frames of images with the same resolution and colour depth. The model is training for $40$ epochs with learning rate decay and ADAM optimiser. To reduce overfitting against factors such as global drift of pixel intensity or cell drift, we adopt the same data augmentation methods used in \cite{GallusserSelfsupervisedDense2023} which includes flipping the order of each crop pair with equal probability, arbitrary rotations and elastic transformations jointly for the pair, translations for each crop in the pair, spatial scaling, additive Gaussian noise, and intensity shifting and scaling. We show various training performance metrics in Figure \ref{fig:TAP pretraining metrics}.

\newpage
\section{Model architecture for the ResNet head used in our experiments}
\begin{figure}[!h]
		\centering
		%
		%
		\begin{subfigure}[h]{0.40\textwidth}
			\includegraphics[width=\linewidth]{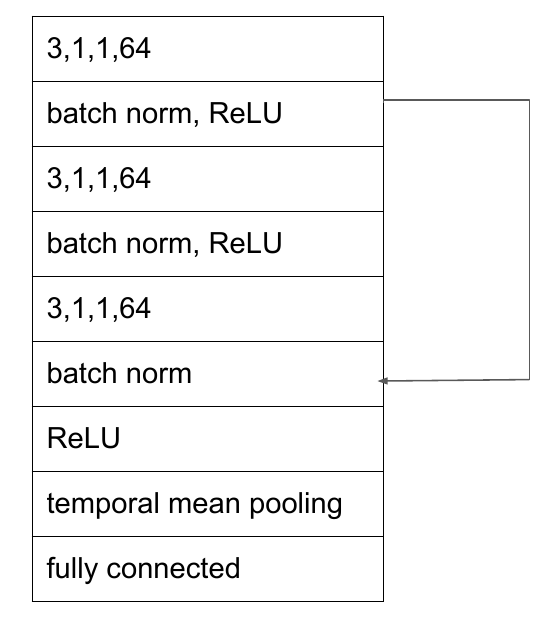}
		\end{subfigure}
		\vspace{1mm}
		\caption{Architecture of the simple ResNet used as the classification head our work. The first row represents the first layer in the network. Numbers in the first row (and similarly in the third and the fifth) refer to the kernel size, stride, padding and the output channel of the corresponding convolutional layer. Temporal mean pooling refer to mean pooling across the temporal dimension. The arrow denotes the shortcut connection.}
		\label{fig:resnet_head_architecture}
\end{figure}

We present the architecture of the ResNet used as the classification head in our model in Figure \ref{fig:resnet_head_architecture}.
\newpage
\section{Evaluating TAP using different crop sizes and resolutions}\label{supp sec:tap accuracy}
We are interested in how the choice of crop size and resolutions of the image where the crops are taken from will affect the validation accuracy of TAP. Due to limited GPU memories we have, we were not able to change the crop size when we train the model for TAP. However, we were able to change the crop sizes during validation. We hypothesise that the TAP accuracy is influenced mainly by the content view by which we mean the scale of the visual information contained in the crop. An analogy is the area a map covers when we vary the scale. Enlarging the content view should increase the TAP accuracy as it allows more cells to appear in the crop which could provide more signals for the model to predict the time arrow. We consider two ways of achieving a larger content view: 1: increasing the crop size, 2: reducing the resolutions (i.e. downsizing) while keeping the crop size. Results are presented in Table \ref{tbl:content_view_TAP_accuracy}.
\begin{table}
    \centering
    \scalebox{0.7}{
    \begin{tabular}{cccccc}
    \toprule
     & same crop size & halved crop size & doubled resolution & doubled crop size & halved resolution\\ 
    \midrule
     & $0.974 \pm 0.001$ & $0.824 \pm 0.002$ & $0.709 \pm 0.001$ & $1.0 \pm 0$ & $0.938 \pm 0.938$ \\ 
    \bottomrule
    \end{tabular}
    }
    \caption{Accuracies of time arrow prediction. In each setting, we run the validation 3 times and report their mean and standard deviation. The resolution of the original image is 1024*1024 and the crop size using in training is 96*96. Halved crop size is 48*48 and halved resolution is 512*512. Doubling works in a similar way. Validation is performed on cell movies obtained from a different location in the same experiment. We can observe that decreasing the content view by either halving the crop size or doubling the resolution while keeping the crop size will reduce the accuracy. Vice versa. Notice that reducing the resolution might not further increase the accuracy.}
    \label{tbl:content_view_TAP_accuracy}
\end{table}
We found that either of these two methods improved the validation accuracy but the increase using 1 is bigger than 2. We think that the reduction in image quality when downsizing might have contributed to the decrease in the accuracy gain. So how should we decide on the crop sizes in cell event recognition using TAP features? We think that one needs to consider and balance the following factors: computing resources (i.e. GPU memories), accuracy of the time arrow prediction and the desired level of accuracy of the cell event recognition. 
\end{document}